# Extract and Merge: Merging extracted humans from different images utilizing Mask R-CNN

Asati Minkesh[1], Kraisittipong Worranitta[2], Miyachi Taizo[3]

Tokai University, Kanagawa, Japan
`minkeshasati@gmail.com`[1], `worranitta.mink@gmail.com`[2],
`miyachi@keyaki.cc.u-tokai.ac.jp`[3]

**Abstract.** Selecting human objects out of the various type of objects in images and merging them with other scenes is manual and day-to-day work for photo editors. Although recently Adobe Photoshop released "select subject" tool which automatically selects the foreground object in an image, it still requires fine manual tweaking separately. In this work, we proposed an application utilizing Mask R-CNN (for object detection and mask segmentation) that can extract human instances from multiple images and merge them with a new background. This application does not add any overhead to Mask R-CNN, running at five frames per second. It can extract human instances from any number of images or videos for merging them together. We also structured the code to accept videos of different lengths as input and length of the output-video will be equal to the longest input-video. We wanted to create a simple yet effective application that can serve as a base for photo editing and do most time-consuming work automatically, so, editors can focus more on the design part. Other application could be to group people together in a single picture with a new background from different images which could not physically be togethcer. We are showing single-person and multi-person extraction and placement in two different backgrounds. Also, we are showing a video example with single-person extraction.

**Keywords:** human extraction; extract and merge; Mask R-CNN; person detection; instance segmentation

## 1      Introduction

Computer vision is a computer science field that seeks to develop techniques to improve the understanding of computer-software from the content of digital images such as photographs and videos. In order to understand the images, computers need to see an image and understand the content that the image contains. To achieve that, computer-software should have the capability to describe the images or videos content, summarize content, and also recognize faces that appeared in those images or videos. To understand the content might also involve extracting a description of those images which it could be an object, text, position, and so on. The goal of computer vision is to make computers understand a digital image. Even though the recent progress on computer vision is still not enough to solve these fundamental problems but with new technologies and research, it brings humanity closer to the goal of computer vision.

In the related research of computer vision, human behavior analysis has a unique position. Extracting a person from the video is a fascinating task for image segmentation but to accurately masking an object requires much training the masked dataset; also, it takes more time in training as the dataset grows.

In our application, we utilize an instance segmentation algorithm proposed in Mask R-CNN, which is an extension of Faster R-CNN (object detection) that masks specific pixel on the object of interest in the image.

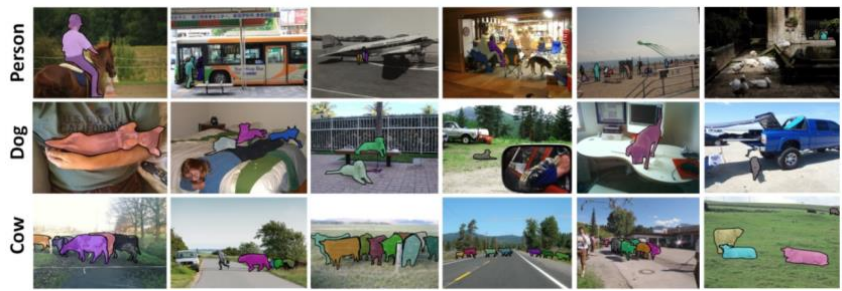

**Fig. 1.** Samples of annotated images in the MS COCO dataset from [12].

We used pre-trained network weights of Mask R-CNN, which was trained on MS COCO dataset (Fig. 1). So, here we are discussing it briefly. In 2014, T.Y. Lin et al. introduced Microsoft Common Objects in context or MS COCO dataset [12]. It is a large-scale dataset that contains multiple objects in everyday scenes. Comparing with ImageNet dataset [13], COCO has more labeled objects per category even it has fewer categories than ImageNet. The object in COCO dataset labeled using per-instance segmentation for more precise in object localization. COCO dataset contains 91 common object categories, and total dataset has 2,500,000 labeled instances in 328,000 images. The goal of COCO is to improve the computer vision tasks for scene understanding and to advance the state-of-the-art in object recognition.

In this work, we use Mask-RCNN algorithm for detecting and generating the mask of the person in a picture, and we are contributing an algorithm for extracting the detected person (all or some of them) from different images and then placing them together with a new background. Nowadays, extracting the person and merging them together as a single image is still a manual work in photo editing, and it becomes even harder with video editing. We would like to have a quick and accurate artificial intelligent for that work so everybody could use it for more further purposed.

## 2 Related work

### 2.1 Instance segmentation application

There are many approaches for instance segmentation that is based on segment proposals. For example, Li et al. [1] combine the segment proposal [4] with an object detection system [2] for fully convolutional instance-aware semantic segmentation or FCIS [1]. However, FCIS still have a problem on the overlapped instances comparing to Mask R-CNN in Fig.2. Because instead of segment proposal, Mask R-CNN based on a parallel prediction of masks and class labels.

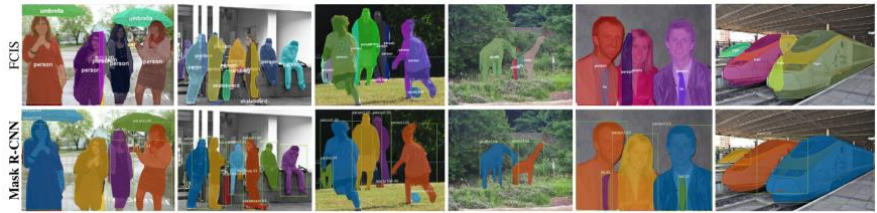

**Fig. 2.** FCIS+++ [1] exhibits spurious edge on overlapping objects comparing with Mask R-CNN with ResNet-101-FPN. [3]

Furthermore, instance segmentation with Mask R-CNN application has appeared in various field. Rajaram Anantharaman et al. proposed the application of utilizing Mask R-CNN for detection and segmentation of common oral diseases [16]. Not only oral pathology, but Mask R-CNN also found in another research that related to the medical industry. Recently, Xiaoyu Tan et al. proposed a Robot-Assisted training in laparoscopy to improve the experiences of surgeons [17]. The Robot-Assisted laparoscopy training system used Mask R-CNN to perform the semantic segmentation of surgical tools and targets for enhancing the automation of training feedback, visualization, and error validation. The static analysis shows that utilizing the training system; trainees skill statistically improves.

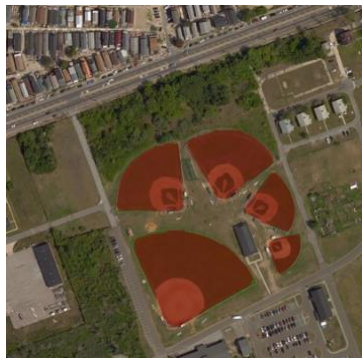

**Fig. 3.** Mask R-CNN finding the baseball field in [21]

Alternatively, using data collected from Google Earth, Shanlan Nie et al. introduced the application for inshore ship detection based on Mask R-CNN [18]. Another approach is to improve OpenStreetMap using Mask R-CNN to detect a feature in satellite images and adding sports fields on OpenStreetMap [21]. These two applications are combined the deep learning technology with remote sensing images which could be useful for remote sensing application in future.

However, there are very few works we found that are closely related to our application. Hiukim Yuen experiments on Mask R-CNN deep learning image segmentation techniques [19]. The inspiration of Hiukim Yuen is his client consulted about the image blending project, so he implemented the solution that allows users to take a photo of themselves and blend it with historical photos. The challenging of this experiment is taken photos are from end-users, which could be any background and to extract and blend users in the photo is quite tricky. Utilizing Mask R-CNN for capturing the persons, he extracts the persons by their relevant pixels and merges them into the target historical photos. After extract the persons, he dilates the mask section using OpenCV dilation API to make the person more contrast and tune the color to black-and-white to match with historic image color tone. The result of image segmentation blending is imposing and could be used in other image processing or behavior understanding.

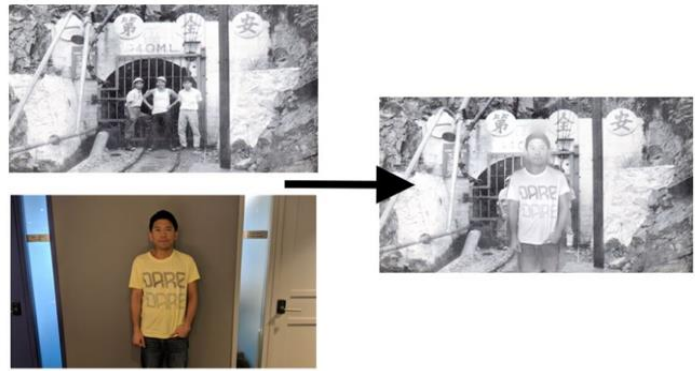

**Fig. 4.** The result of extract and blend to the historical photo in [19]

Not only extract and blending people, but another related work is segmenting some part of the human for fashion approach. The idea of this work is to take a raw image, segment the article of clothing, and match with the database to find similar accessories. Michael Sugimura built a custom multi-class image segmentation model to classify bag, boot, and top from an image [20]. This application phases start with object detection and then comparisons of input objects in an image to a known database for matching. There are two models in this work, first is object detection to localized objects, and second is a comparison based on localized objects.

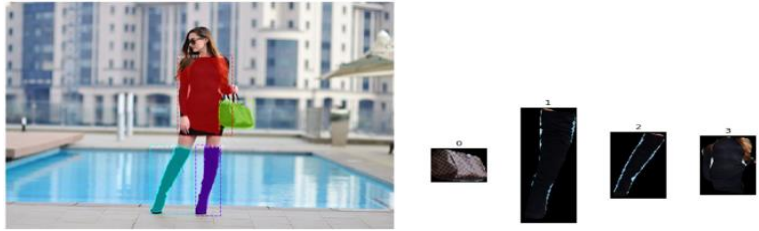

**Fig. 5.** The classification result from [20]

Given 100 training images, this work trains the Mask R-CNN model using pre-trained weights from the COCO challenge. This application use pixel-level segmentation instead of bounding box because of the cleaner output of object detection make the comparison stage easier and more optimized. Fig. 5 shows the example of the object detection and also the extraction part of top, boot and bag.

With the incredible related work above, it proved that image segmentation application and deep learning could uncover new insights and even more advanced application.

### 2.2　Adobe Photoshop Tools

In Adobe Photoshop, many tools have been used for extracting person in a photo as we describe some of it below:

**Select Subject:** Select subject is an edge-detection based tool from Adobe Photoshop. Select subject automatically selects the foreground instances in an image but we must refine the selection manually to get an accurate selection.

**Magic Wand Tool:** It is known as Magic Wand. Magic Wand is a very famous tool that has been in Adobe Photoshop for a very long time. The basic idea of the Magic Wand is to select pixel based on tone and color, which is different from the select-subject that is based on edge-detection.

## 3　Problem Definition

We want to point out some limitations of the related work in section 2. From section 2.4, tools from Adobe Photoshop (Select Subject) might be useful in extracting person, but for merging, it still requires a lot of manual actions, and it mainly useful when there is only one object in the image, but in case of multiple objects, it doesn't work and most of the work need to be done manually because Select Subject use edge detection in the background, so, when it comes to an elaborate scene or crowded scene where objects are coupled together, its detection results are quite random. Unlike Select Subject, our application does segmentation pixel-by-pixel, so it is more precise, and It can perform the extraction also it merges the extracted objects together automatically layer by layer. So, the user does not need to do the manual part for extraction and do not need to deal with the layers for merging.

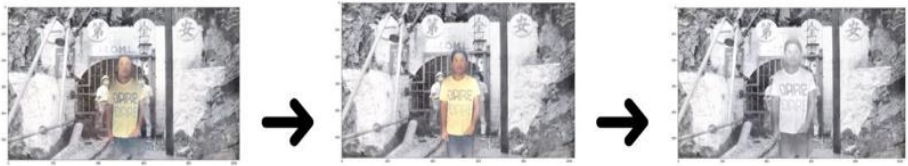

**Fig. 6.** Extract and blending on [19]

Although we found one application [19] (medium post) that is very close to our application that could extract human-object and blend them with a historical place's image, as the process shows in Fig.6, but our approach is entirely different because our objective is to merge human-objects after extracting them from different images that is a very crucial task in photo editing. Moreover, our application outperforms in various aspects such as it could not extract human objects from multiple inputs, and also it has the limitation only for a photograph, do not work with videos. The major problem in this work is the first result came out blurry and need to sharp it with other tools. After sharping the human instances, it also needs to tone the color to black-and-white shows in Fig.6. The Extract and Blend required many processes of work, so it is not easy to run.

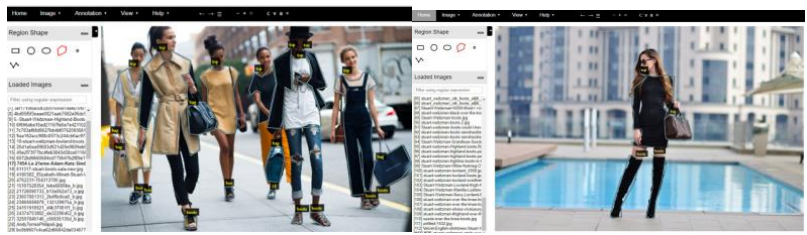

**Fig. 7.** Data annotation from [20]

In this work [20], their idea is to segment the article of clothing and match with the database to find similar items. Fig.7 shows data annotation of this work. It required a data annotation to label the data class, which are top, shoes, and bag. Labeling all of the information in the images is entirely a manual work which requires a lot of time and effort. Concerning the works described above, our technique for extract and merge is more straightforward. Based on the area of the target person, and our application can extract and merge as many people (extracted from different images) as user required.

Selecting (Extracting) image instances from different images and merge them together is a frequent and crucial work for the Photo Editors, and it becomes even harder with the video. We wanted to create a simple way to do that using artificial intelligent approach. Also, this could be used for travel agency advertising because making the customer seeing themselves in the landmark or beautiful places, that kind of image bringing the imagination into vision.

## 4 Proposed Method

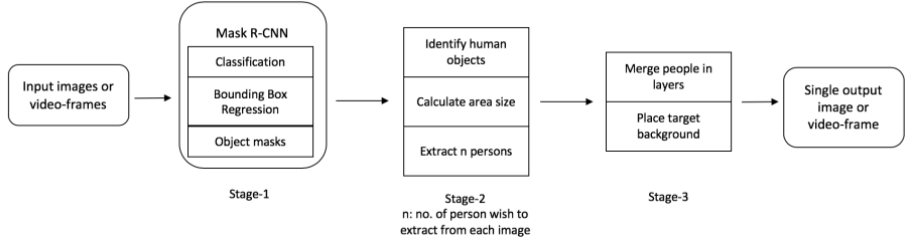

**Fig. 8.** Purposed method diagram

In this paper, we proposed a merger application that detects and extracts only human-object from 2 or more images then place them together with a new background. As an input, the application takes 2 or more images consisting various objects (e.g., human, animal, bikes) and a background image, as an output, we get a single image with target background consisting only human object extracted from different input images.

Since our application uses Mask R-CNN for instance segmentation and mask generation, and, Mask R-CNN is based on Faster R-CNN architecture. So, we begin by briefly reviewing the Faster R-CNN[9] and Mask R-CNN[3]. Faster R-CNN consists of two stages. The first stage called a Region Proposal Network (RPN), proposes candidate object bounding boxes. The second stage, which is in essence, Fast R-CNN [14], extracts features using RoIPool from each candidate box and performs classification and bounding-box regression. The features used by both stages can be shared for faster inference. Faster R-CNN has two outputs for each candidate object, a class label, and a bounding-box offset; to this Mask R-CNN added a third branch that outputs the object mask —which is a binary mask that indicates the pixels where the object is in the bounding box. However, the additional mask output is distinct from the class and box outputs, requiring extraction of the much more exquisite spatial layout of an object. To do this Mask RCNN uses the Fully Convolution Network (FCN).

Now, we explain our end to end application structure. Application takes two inputs and gives one output. The first input is, n (n>1) number of images contains various type of objects and scenes, second input is, a background image, and output is, a single image in which all extracted human objects (from input images) placed together into target background (second input image). Our application has three stages.

In the first stage, we do instance segmentation and generate their mask for each detected object in each input image using Mask R-CNN. In the second stage, we identify the human-object out of all type of objects and then select an expected number of human-object (based on their area size) which we need to extract for each input image separately. In the third stage, we extract the selected person (in the second stage) and place them into the target background image one by one (layer by layer).

Here, we explain the three stages of our application in detail. Before feeding the input images into the first stage, we resized all input images and background image to the same size. The first stage is pretty straight forward because here we are using Mask R-CNN without any changes in their architecture for object detection (bounding box offset), object classification (object name), and mask generation (pixels which belong to the object). In the second stage, we calculate the area of all detected human-object by using the bounding box offsets ( Y1 , X1 , Y2 , X2 ) with the following formula.

$$\text{Area} = (Y2 - Y1) * (X2 - X1) \quad [\text{Rectangle Area} = \text{Height} * \text{Length}] \quad (1)$$

We store this area with respective person ids in descending order, in order to be able to extract main people who are significant in the image, and, select the people to extract (if we expect to extract n person from each image then consider first n person ids). We repeat this entire second stage for each input image. In the third stage, since our all input images, background image, and generated masks are of the same size that's why we can compare pixel-by-pixel and replace the pixels of the background image with input images pixels where we identified a human object. So, for each selected human-object in an input image, we replace the background image pixel-value with the image pixel-value where the mask pixel-value is true (which pixels belongs to the object). Moreover, we do this layer by layer. We repeat this entire third stage for each input image. Here, we can see that people of the last input image will be in the top layer and most visible if objects are on the top of each other.

## 5 Implementation and Results

### 5.1 Hardware Environment

In this project, we used an i5 processor (6-cores), 32-GB RAM, 3-TB ROM, and Nvidia Titan V Graphics card of 12-GB memory. We were able to process around 3-frames per second when 1-image were feeding into the net in parallel, and 5-frames per second when 2-images were feeding into the net in parallel.

### 5.2 Implementation

In the first stage of our implementation, we used the open-source implementation of Mask R-CNN using Python, Keras, and TensorFlow [15]. It is based on Feature pyramid network (FPN) and Resnet101 backbone. We used pre-trained weights of this network that was trained on MS COCO dataset. That network was trained on 62 categories of objects. Most of this implementation follows Mask R-CNN, but there are some cases where they deviated in favor of code simplicity and generalization.

There are mainly 3 differences. First is, they resized all images to the same size to support training multiple images per batch. Second is, they ignored bounding boxes that come with the dataset and generate them on the fly to support training on multiple datasets because some dataset provides bounding boxes and some provide masks only. This also made it easy to apply image augmentation that would be harder to apply to the bounding boxes, such as image rotation. The third is, they used a lower learning rate instead of using 0.02 that was used in the original paper. Because they found it to be too high, that often causes the weights to explode, especially when using a small batch size. They related it to differences between how Caffe (original paper implemented in Caffe) and TensorFlow computes gradients (sum vs. mean across batches and GPUs). Alternatively, maybe the official model uses gradient clipping to avoid this issue. They also used gradient clipping but did not set it too aggressively.

Second stage and third stage are implemented using Python and its libraries for extracting and merging the extracted objects. We structured the code in a way to be able to extract single or multiple people out of all detected human objects from an image. To handle any number of inputs (images or videos), we placed the extracted objects into a background image layer by layer (as we do in photoshop to have a clean look) to avoid any mixing and human object's edges will be visible even though objects are on the top of each other. In the case of video inputs, output video-length is a crucial aspect because if the length of input-videos is different then how to figure out the length of output-video. To solve this, we iterated our main application on frames until the frames of longest input-video have finished.

### 5.3 Results

We are showing four types of results.

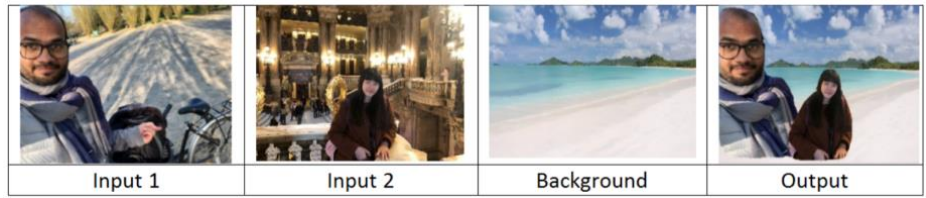

**Fig. 9.** Example of 2-images input, extracted one person from each image.

In Fig. 9, there are two input images captured in wild and extracting one human object from each and placed them together with a beach-background. We have placed a man from Input-1 in layer-1 and a woman from input-2 in layer-2.

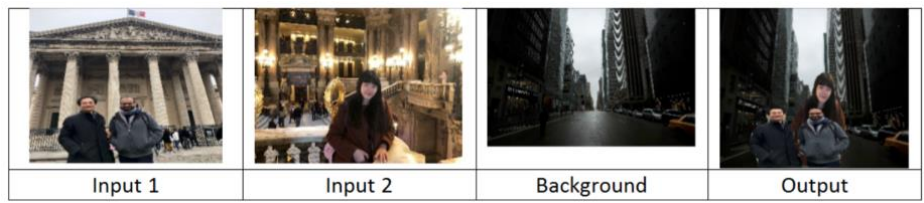

**Fig. 10.** Example of 2-images input, extracted two-person from one image, and one person from another.

In Fig. 10, there are two input images captured in wild and extracting two people from 1st input image and one person from 2nd input image. Here, a woman from input-2 is in the bottom layer and two-man from input-1 is on top-layer.

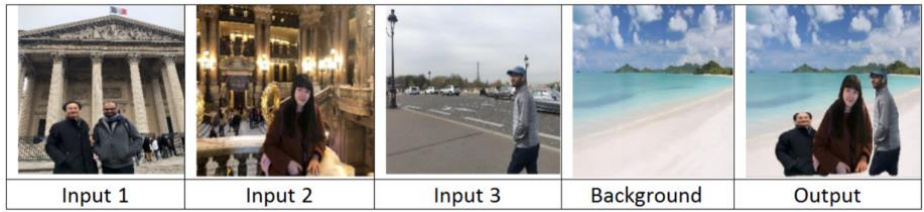

**Fig. 11.** Example of 3-images input, extracted one person from each image.

In Fig. 11, there are three input images captured in wild and extracting one person from each. A man standing in left has a bigger area size as compared to the other man in the Input-1, so, we when we extracted one person from each image then left only left side person was extracted.

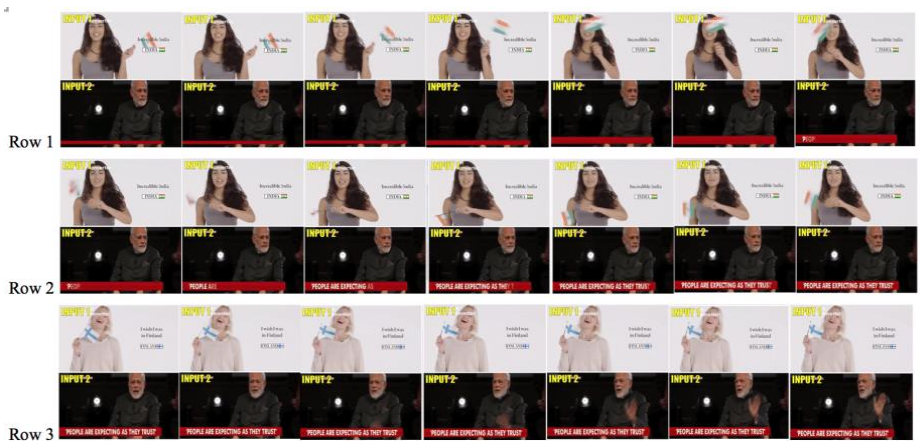

**Fig. 12.** Total of 21 images in 3-rows. Each image consists of 2 frames (one frame of each input video).

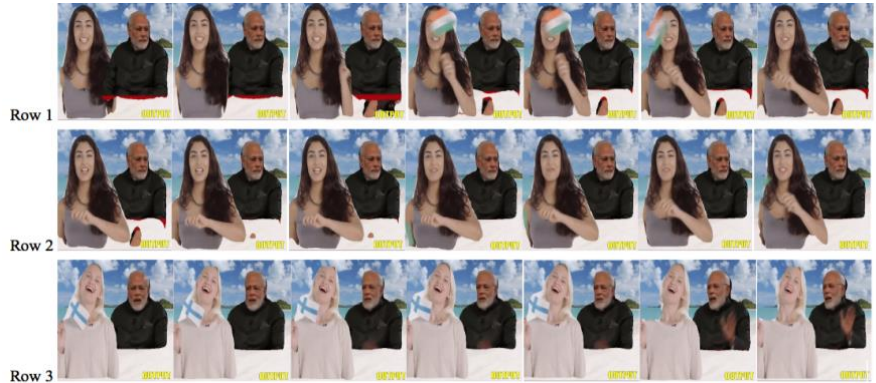

**Fig. 13.** 21 frames of one output video.

In Fig. 12 and 13, It is shown the merging and extraction of a single person from 2 input videos (one person from each) frame by frame. In Fig. 12, there are 21 cells in 3-rows. Each cell consists of 2 frames (one frame of each input video) and Fig. 13, there are also 21 cells, and each cell consists of one frame of output video. Each cell of Fig. 12 is showing the result of its corresponding cell in Fig. 13.

## 6       Conclusion

In this paper, we purposed an automatic way to extract persons from different images and merge them together as a single image with a new background. We structured the code in a way to be able to extract single or multiple people out of all detected human objects from an image. It can handle any number of input images or videos and it places the extracted objects into a new background layer by layer to avoid any mixing. Because our application is dependent on Mask R-CNN for object detection and mask generation, that is why we could not improve the accuracy in human detection. Even though It does not make any overhead to Mask R-CNN and running at 5 frames per second, but it is not sufficient for real-time application. However, to improve the accuracy or speed could be done by designing the new architectures for masking only the human objects in images instead of generating masks for all type of objects and then consider only human-objects to extract.

## 7       Acknowledgment

We want to thank Japan International Cooperation Agency for the funding support under the Innovative Asia Program 2017.

# 8 References

1. Y. Li, H. Qi, J. Dai, X. Ji, and Y. Wei. "Fully convolutional instance-aware semantic segmentation," In CVPR, 2017.
2. J. Dai, Y. Li, K. He and J. Sun. "R-FCN: Object detection via region-based fully convolutional networks," In NIPS, 2016.
3. K. He, G. Gkioxari, P. Dollar and R. Girshick. "Mask-RCNN," In ICCV, 2017.
4. J. Dai, K. He, Y. Li, S. Ren, and J. Sun. "Instance-sensitive fully convolutional networks," In ECCV, 2016.
5. J. R. R. Uijlings, K. E. A. van de Sande, T. Gevers and A. W. M. Smeulders. "Selective Search for Object Recognition. International Journal of Computer Vision," In IJCV, 2013.
6. R. Girshick, J. Donahue, T. Darrell, and J. Malik. "Rich feature hierarchies for accurate object detection and semantic segmentation," In CVPR, 2014.
7. Y. LeCun, L. Bottou, Y. Bengio and P. Haffner "Gradient-Based Learning Applied to Document Recognition," In Proc. of the IEEE, 1998s
8. A. Krizhevsky, I. Sutskever, and G. Hinton. "ImageNet classification with deep convolutional neural networks," In NIPS, 2012.
9. S. Ren, K. He, R. Girshick, and J. Sun. "Faster R-CNN: Towards Real- Time Object Detection with Region Proposal Networks," In NIPS, 2015.
10. J. Redmon, S. Divvala, R. Girshick and A. Farhadi. "You Only Look Once: Unified, Real-Time Object Detection," In CVPR, 2016.
11. W. Liu, D. Anguelov, D. Erhan, C. Szegedy, S. Reed, C.-Y. Fu and A. C. Berg. "SSD: Single Shot MultiBox Detector," In ECCV, 2016.
12. T. Lin, M. Maire, S. Belongie, J. Hays, P. Perona, D. Ramanan, P. Dollar, C. L. Zitnick. "Microsoft COCO: Common Objects in Context," In ECCV, Part V, LNCS 8693, pp. 740–755, 2014.
13. J. Deng, W. Dong, R. Socher, L.-J. Li, K. Li and L. Fei-Fei. "ImageNet: A Large- Scale Hierarchical Image Database," In CVPR, 2009.
14. R. Girshick. "Fast R-CNN," In ICCV, 2015.
15. Matterport Inc. "Mask R-CNN for object detection and instance segmentation on Keras and TensorFlow," Retrieved 2017.From https://github.com/matterport/Mask_RCNN
16. R. Anantharaman, M. Velazquez and Y. Lee "Utilizing Mask R-CNN for Detection and Segmentation of Oral Diseases," In BIBM, 2018
17. X. Tan, C.-B. Chng, Y. Su, K.-B. Lim and C.-K. Chui. "Robot-Assisted Training in Laparoscopy Using Deep Reinforcement Learning", In IEEE Robotics and Automation Letters 4, 2019.
18. S. Nie, Z. Jiang, H. Zhang, B. Cai and Y. Yao. "Inshore Ship Detection Based on Mask R-CNN," In IGARSS, 2018.
19. HiuKim Yuen. "Image blending with Mask R-CNN and OpenCV," Retrieved July 22, 2018 from https://medium.com/softmind-engineering/image-blending-with-mask-r-cnn-and-opencv-eb5ac521f920
20. Michael Sugimura. "Stuart Weitzman Boots, Designer Bags, and Outfits with Mask R-CNN," Retrieved October 8, 2018 from https://towardsdatascience.com/stuart-weitzman-boots-designer-bags-and-outfits-with-mask-r-cnn-92a267a02819
21. Jason Remillard. "Image to OSM," Retrieved February 3,2018. from https://github.com/jremillard/images-to-osm